\documentclass[largeformat]{interact}

\usepackage{epstopdf}
\usepackage{graphicx}%
\usepackage{multirow}%
\usepackage{amsmath,amssymb,amsfonts}%
\usepackage{amsthm}%
\usepackage{mathrsfs}%
\usepackage[title]{appendix}%
\usepackage{xcolor}%
\usepackage{textcomp}%
\usepackage{manyfoot}%
\usepackage{booktabs}%
\usepackage{algorithm}%
\usepackage{algorithmicx}%
\usepackage{algpseudocode}%
\usepackage{listings}%
\usepackage{url}
\usepackage{caption}
\usepackage{subcaption}
\usepackage{siunitx}
\usepackage{wrapfig}
\usepackage{float}
\usepackage{comment}
\usepackage[]{natbib}

\theoremstyle{plain}

\theoremstyle{definition}

\theoremstyle{remark}

\begin{document}

\articletype{ARTICLE}

\title{ARTICLE PUBLISHED IN INTERNATIONAL JOURNAL OF COMPUTER INTEGRATED MANUFACTURING, TAYLOR \& FRANCIS, \url{https://www.tandfonline.com/doi/full/10.1080/0951192X.2024.2421313}}

\subtitle{Smart Automation in Luxury Leather Shoe Polishing: A Human Centric Robotic Approach}

\author{
\name{Matteo Forlini\textsuperscript{a}\thanks{CONTACT Matteo Forlini Email: m.forlini@pm.univpm.it}, Marianna Ciccarelli\textsuperscript{a}, Luca Carbonari\textsuperscript{a}, Alessandra Papetti\textsuperscript{a}, Giacomo Palmieri\textsuperscript{a}.}
\affil{\textsuperscript{a}DIISM— Department of Industrial Engineering and Mathematical Sciences, Polytechnic University of Marche, Via Brecce Bianche, Ancona, 60131, Italy
}}

\maketitle

\begin{abstract}
The polishing of luxury leather shoes is a delicate, labor-intensive process traditionally performed by skilled craftsmen. Footwear companies aim to automate parts of this process to enhance quality, productivity, and operator well-being, but the unique nature of luxury shoe production presents challenges. This paper introduces a solution involving a collaborative robotic cell to assist in shoe polishing. A collaborative robotic manipulator, equipped with a specialized tool and governed by force control, executes the polishing tasks. Key factors such as trajectory design, applied force, polishing speed, and polish amount were analyzed. Polishing trajectories are designed using CAM software and transferred to the robot’s control system. Human operators design the process, supervise the robot, and perform final finishing, ensuring their expertise is integral to achieving quality. Extensive testing on various shoe models showed significant improvements in quality and reliability, leading to successful implementation on an industrial production line.
\end{abstract}

\begin{keywords}
Shoe Polishing, Collaborative Robotics, Human Computer Interaction, Human Centred Manufacturing, Human Robot Collaboration
\end{keywords}

\section{Introduction}
\begin{figure}[h]
\centering\includegraphics[width=1\linewidth]{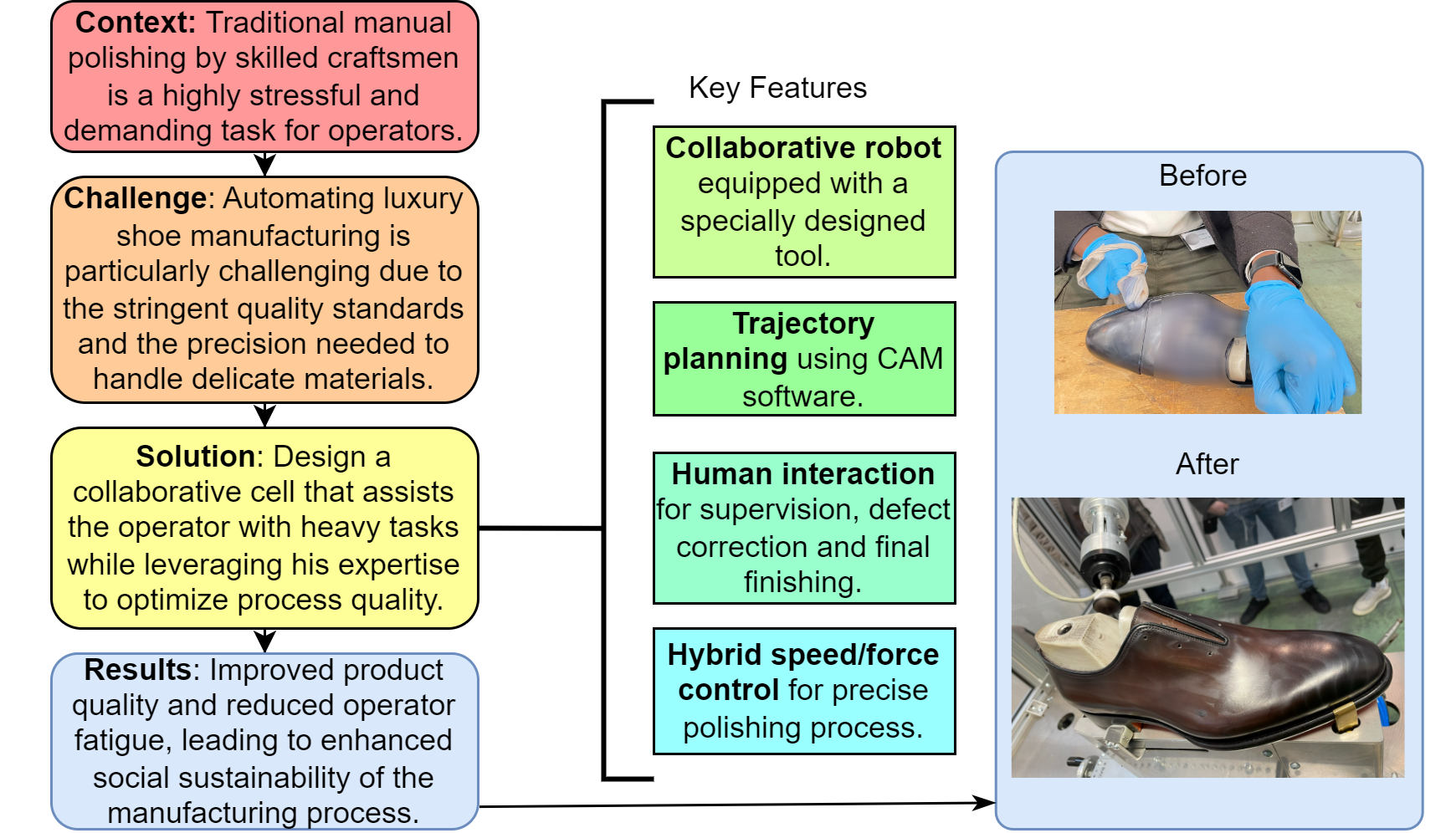}
\caption{Research problem and applied methods.}
\label{fig:graph_abstrac}
\end{figure}

Nowadays, many manufacturing companies, in order to be more competitive in the global market, need to automate their process to achieve better results on product quality and productivity.
Not only large companies, but also small ones are moving toward process automation, due to the need to improve mental and physical well-being of the workers. Human-centred manufacturing is crucial to increase flexibility, agility and competitiveness \citep{lu2022outlook}. Humans are no longer meant to do tedious and strenuous work, but to add high-value content to final products by exploiting their cognitive abilities. The relationship between humans and machines has radically changed with the transformation of Industry 5.0: machines are required to perceive and adapt to workers' needs, not vice versa. Production is changing in all companies in different sectors, even in those that have production characteristics totally opposed to an automated process. For example, in footwear luxury industries the processes are typically manually performed by expert artisans due to high quality, precision and comfort requirements of products and to the wide variability and customisation of production. All these features are in contrast to full automation, and only craft work can bring profitability to the company. However, the need for automation is strong to ensure high productivity but also the well-being of the workers. A central process in the production of shoes is polishing. In the luxury sector this process is carried out at present by skilled craftsmen that are able to ensure the best quality of the product and the right shine and color to the leather. The variability of the polishing process is very wide, depending on different factors such as leather color, leather type, shoe model and so on. Unfortunately this process is also one of the most tiring because it requires a long time (about 20-23 minutes per shoe) with tedious and repetitive movements. Automation of this process would reduce the physical effort of the operator, who could simply supervise the operation and control the quality of polishing, improving it if the automated process did not meet the requirements.
A collaborative robot can be used for this purpose since it can work alongside and cooperating with the operator. In this type of industry there is a reluctance to completely replace the manual process, because it is the human experience that gives the product the highest level of quality in terms of aesthetic beauty, tactile feel, and comfort. Therefore, automation involving human cooperation and supervision is the only one possible to combine the two requirements of ensuring a high-quality product and high human well-being. In Figure \ref{fig:graph_abstrac} is illustrated the research problem with the context and the methodology applied to fill the  gap.

\subsection{Related Works}
Several studies can be found in literature on the automation of the footwear industry. Initial solutions for the automation of different shoe production processes, such as shoe lasting machine, shoe gluing and shoe finishing, were presented in \citet{nemec2006automation}. 
In \citet{nemec2008shoe} a shoe grinding robotic cell was developed using the robotic kinematic redundancy due to the circular shape of the grinding disc. Also roughing and cementing are important processes that can be automated \citep{jatta2004roughing, hu2007automatic, pedrocchi2015design}. In \citet{choi2008development} a new robot with a deburring tool is studied and implemented for shoe buffing and roughing. \citet{kim2022robot} developed a robotic work cell with an industrial robot for upper roughing and cementing and for the sole cementing with a system for path tool planning. The problem of innovation and automation in footwear industries is well known also in the European Commission where different projects were founded in the past. The ROBOFOOT consortium was founded in 2010 with the aim of promoting the introduction of robotics in the European Footwear Manufacturing Industry \citep{maurtua2012robotic}. The consortium remarks the importance of the coexistence of manual operations with automated ones. In the  European IDEA-FOOT project a new method for the shoe integrated design and production was introduced: most of the production parameters were derived from the shoe 3D CAD model and an innovative automated production plant in which manipulators and automated machines are fully integrated was developed \citep{cocuzza2013novel}. Some studies have been also conducted on the shoe gluing: in \citet{castelli2020feasibility} a robotic cell for glue deposition on shoe uppers based on an extrusion system for the deposition of molten material originally in the form of a filament was analyzed and proposed. Two solutions were tested: in the first, the extruder is fixed while the robot moves the upper under the extruder; in the second, on the contrary, the extruder is fixed to the robot and moves over the upper to deposit glue. The force exerted by the robot on the extruder is a crucial parameter to have a correct glue deposition; for this reason a load cell was mounted on the robot flange. 
Some robotics application for footwear gluing were developed using computer vision to plan the tool path \citep{pagano2020vision, lee2018implementation}. The vision system enables the location and reconstruction of the shape of objects to be glued; the detected information is then used to plan the trajectory of the robot. An important aspect concerns also with the material handling for the shoe production: \citet{oliver2021towards} developed a novel sole grasping method, compatible with soles of different shapes,
sizes, and materials. The method computes antipodal grasping points from point cloud acquired by RGB-D cameras or laser, without requiring a previous recognition of sole. In \citet{gracia2017robotic, mendez2020robotic} a shoe packaging robotic system is presented: the robot moves the shoes from the conveyor to the box and at the same time is able to detect human interaction and the dynamic environment, modifying its behavior accordingly. A monitoring of robotics cell for footwear industries according to principles of Industry 4.0 is done in \citet{roman2018distributed}.
Regarding the automation with a robotic cell of the shoe polishing cell some works have been done in literature.
In \citet{nemec2008robotic}, in particular, a computer aided design (CAD) approach for automatic generation, optimisation and validation of motion trajectories was integrated into a robotic cell for shoe polishing. In this scenario, the shoe is attached to the robot end-effector and is then moved against a fixed rotating brush without controlling the force executed.  \citet{borrell2023cooperative} presented an innovative approach for automatic shoe patina growing in the footwear industry using a collaborative robot and a dedicated polishing tool. In this case, the trajectory is taught by manually moving the robot with the tool along the surface of the shoe, following the desired path.

\subsection{Aim of the Research}
The objective of the presented work is to design and develop a collaborative robotic cell for luxury shoe polishing. To the best of the authors' knowledge, there is currently no robotic system described in the literature that can perform a complete polishing process on the entire shoe with results comparable in quality to those achieved by manual polishing. This highlights a research gap between the existing literature and the requirements of this particular application. Briefly, this gap can be outlined in the following aspects: a) control of the amount of polish applied, b) maintaining a constant force applied by the robot, c) ensuring that the tool remains perpendicular to the shoe surface, d) precise control over the types of trajectories performed and e) management of the tool speed during polishing. These features are essential to achieve high-quality results in luxury leather footwear, as will be explained in detail below.\\
In the literature, only one study has explored the use of a collaborative robot for polishing luxury leather shoes \citep{borrell2023cooperative}. This paper will provide a detailed comparison between that study and the proposed approach, emphasizing key differences and novel contributions.
At present, the process is done completely by hand in two stages: a thicker and more viscous polish is used in the first stage, while the second stage is for finishing, to give the right shine and shade to the shoe.
The first step in the process is the one to be automated since it takes the most time and is the most tiring for the operator. Then the operator will perform the final finish with a smoother polish. In addition, the cycle time required by the company must be met.
To achieve a good finish, polishing of the entire shoe should be done by performing different types of trajectories for each area of the shoe. These should start and end from different points, so as to leave no polish marks and achieve good homogenization. Other key aspects are the number of trajectories performed and the contact force between tool and shoe. The latter, in particular, should be as much constant as possible; for this reason a force active control \citep{tian2016modeling} can be implemented exploiting the load cell at the wrist of the collaborative robot. Studies can be found on polishing complex surfaces using robots, for example in \citet{wang2019towards}.
Different types of polishing paths, varying for models and sizes, have been drawn by CAD software working on the 3D model of the shoe. Such paths were then converted into scripts executable by the robot with both a tool speed control in the tangent plane and a force control on the normal direction. 
The Universal Robot's UR5e manipulator have been used to implement the process with the integration of a polishing tool and a polish dispenser designed by the authors for the specific application.
The entire process was tested on numerous shoes of different sizes and validated by the company's chemical laboratory, which carefully evaluated the final quality of the processed shoe.The cell is now operational in the production line; the operator merely supervises the first polishing step, monitoring the robot cycle and composing polishing recipes by determining how many and which trajectories to perform on each area of the shoe. The choice of a collaborative robot cell is motivated
by the operator's need to approach the shoe at any time to monitor the process and intervene if necessary. This allows the polishing parameters to be adjusted in real time and any imperfections to be corrected manually without interrupting the robotic operation. Indeed, it is important that the automatic process is not interrupted because a change in the planned trajectories would cause an uneven polish deposit with often irreversible aesthetic defects. As a consequence, the robot must operate in power and force limiting mode (according to ISO/TS 15066:2016) and the polishing tool must be designed to be safe in the event of a collision with the operator, thus fulfilling the risk analysis for a collaborative application.
In summary, the rest of the paper has been organized as follows: critical aspects to be taken into account in the automation of the shoe polishing process are described in Section \ref{sec:critical aspect}; in Section \ref{sec:methodology}, the methodology used in robotic polishing is explained; the results obtained from the tests are presented in Section \ref{sec:exp test}; finally, a discussion of the main aspects of the work and concluding remarks are given in Section \ref{sec:discussion} and Section \ref{sec:conclusion}, respectively.

\section{Shoe Polishing Process: Critical Aspects} \label{sec:critical aspect}

\begin{figure}[h]
\centering\includegraphics[width=8cm]{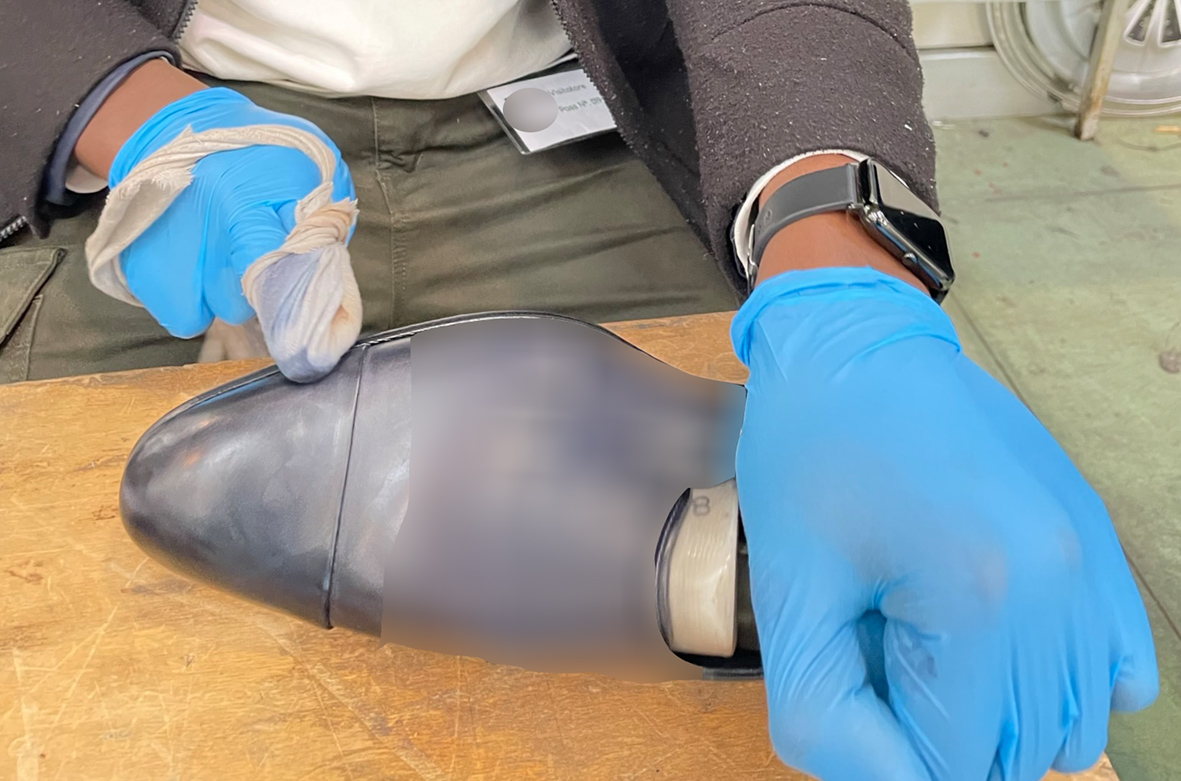}
\caption{Manual operation.}
\label{fig:processo manuale}
\end{figure}

The purpose of this section is to highlight the critical aspects of manual polishing of luxury footwear that were considered in designing the automated process.
The process begins with a careful visual inspection by the operator to understand the type of leather to be treated, the type of polishing cream and amount needed, and whether the shoe will need more work with more polish applications than the standard. The operator takes about always the same amount of polish from a container through a piece of cotton placed on his fingers. He then releases the excess amount of polish onto a pad and begins working on the shoe. Removing the excess and homogenizing the polishing cream left on the piece of cotton is essential to avoid marks on first contact with the leather that would permanently compromise the quality of the shoe. The process starts each time form different points of the upper of the shoe, depending on the desired tone and shading effect. The movements are executed with constant speed and finger pressure on the leather. 
A phase of the manual operation is shown in figure \ref{fig:processo manuale}.

In the second phase, the same operations are repeated with a polish cream that differs for water content.
Step 1 and step 2 are repeated \textit{n} and \textit{m} times, where \textit{n} and \textit{m} vary depending on the type of leather and the finish obtained.

In summary, the crucial factors for the automation of the polishing operation are:

\begin{itemize}
    \item amount of polish cream;
    \item speed of application;
    \item trajectories of the tool;
    \item contact force on the leather surface.
\end{itemize}

How control of the previous process parameters was handled will be described in the following sections. 

\section{Methodology} \label{sec:methodology}
\subsection{Robotic Cell Overview}

A reachability study conducted with ROBODK and Siemens Tecnomatix simulations allowed the type, size, and mounting configuration of the manipulator to be defined. In particular, the Universal Robot UR5e manipulator in an upside-down mounting configuration was identified as the best solution, as it was shown that the workspace free from singular configurations results to be larger then in the traditional mounting configuration. The shoe is fixed on a specially designed support with a jaw lock system actuated by pneumatic valves. The support is mounted on a rotary table, which allows the shoe to rotate of $\ang{180}$ around the vertical axis. In this way all the points of the shoe can be polished keeping the robot in a workspace where the dexterity is high, far from singular configurations.
The tool of the robot, described in detail in \citep{chiriatti2022human}, consists on a spindle with a spring used to compensate for axial forces and a backing pad with a diameter of $30\,\mathrm{mm}$. The speed of rotation was set at $3000\,\mathrm{rpm}$.
The load cell integrated in the UR5e robot has been exploited for force control. The polish dispenser and homogenization stand have been placed on the side of the workbench.
The model and size of the shoe to be processed are entered by the operator through an HMI display. Then, the PLC retrieves the basic polishing recipe from a database, which includes, for example, the type and number of trajectories and the number of polish replenishments; however, the operator can modify the best polishing cycle for that shoe based on his or her experience with simple commands on the HMI touch screen. During the process, thanks to the collaborative nature of the robotic application, the operator can control the process by approaching the shoe and intervening manually if necessary without interrupting the automatic operation. In this way, the operator can profitably play the role of supervisor based on his know-how and experience, which are essential to achieve a process with high final quality. 

\begin{figure*}[htb]
\centering
\includegraphics[width=8cm]{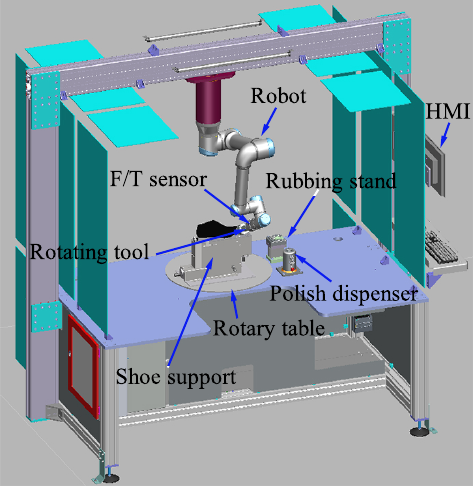}
\caption{Design of the robotic cell.}
\label{fig:banco nx}
\end{figure*}%
\begin{figure*}[htb] 
\centering
\includegraphics[width=10cm]{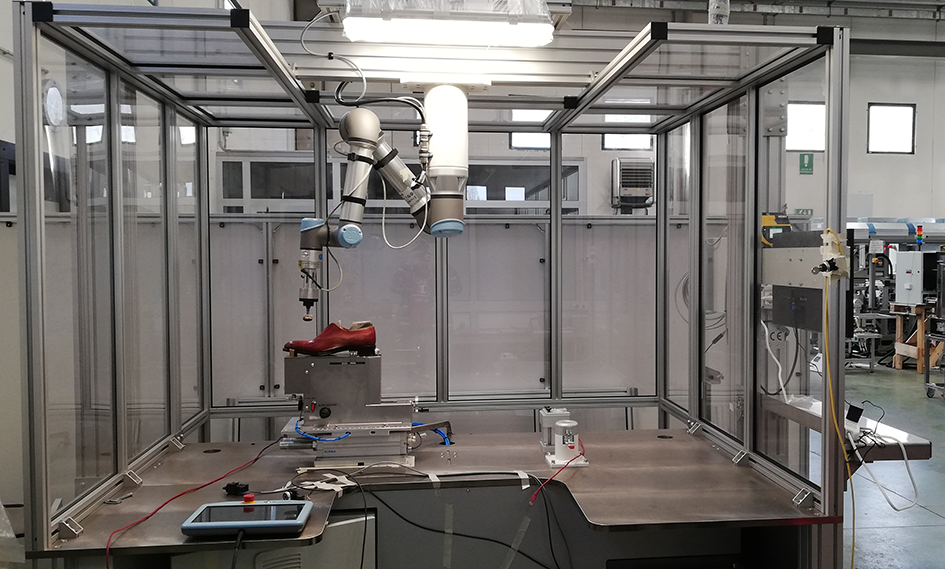}
\caption{The real robotic cell.}
\label{fig:banco reale}
\end{figure*}

Figure \ref{fig:banco nx} shows the design of the workbench with all the components, while Figure \ref{fig:banco reale} shows the real cell. A difference in shoe placement can be seen between the two configurations: after preliminary tests, it was decided to move the heel of the shoe a greater distance from the robot's shoulder to improve the robot's dexterity.

\subsection{Amount of Polish}
\begin{figure*}[htbp]
\centering\includegraphics[width=\textwidth]{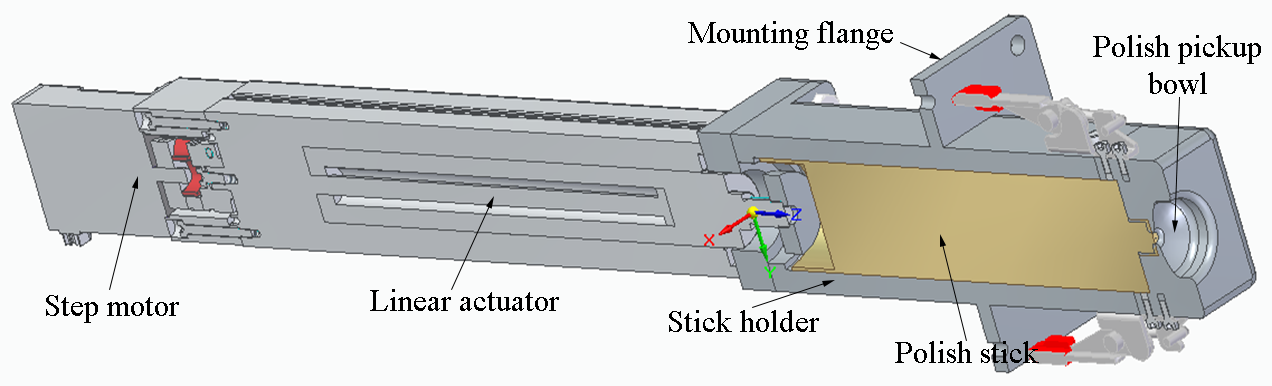}
\caption{3D prospective view of the polishing dispenser with all its components. The step motor drives the linear actuator, which pushes the bottom of the polish stick inserted into the stick holder. In that way the polish is dispensed in a precise amount in the polish pickup bowl.}
\label{fig:dispenser}
\end{figure*}

To ensure a repeatable and controllable process, a predefined and constant amount of polishing cream must be taken at each application. To solve this problem, a dispenser has been designed that dispenses a well-defined amount of polishing cream into a container from which the tool will later make the pickup. The design of the dispenser is shown in Figure \ref{fig:dispenser}. The polish is supplied inside a standard silicone stick with a diameter of $D=50\,\mathrm{mm}$ and a length of $L=200\,\mathrm{mm}$.The stick is inserted into a specially designed stick holder. A stepper motor drives a linear actuator that acts as a piston for the polish stick, which will dispense the polish into the bowl. The pitch of the linear actuator is $p=5\,\mathrm{mm}$ per turn, while the step of the motor is $\theta=\ang{1.8}$; thus, in order to dispense a volume $V_0$ of polish the number $n$ of motor steps can be found as:

\begin{equation}
   n=\frac{1440}{\pi D^2 \theta p} V_0
\end{equation}

\begin{figure*}[htb]
\begin{subfigure}{0.5\textwidth} 
\centering{
\includegraphics[width=5cm]{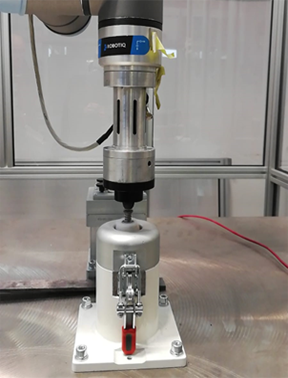}
}
\subcaption{}
\label{fig:prelievo}
\end{subfigure}%
\begin{subfigure}{0.5\textwidth} 
\centering{
\includegraphics[width=5cm]{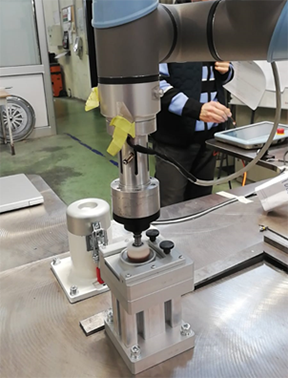}
\subcaption{}
\label{fig:omogenizz}
}\end{subfigure}%
\caption{Polish pick up (a) and homogenization (b).}
\label{fig:subfig prelievo e omog}
\end{figure*}

A FESTO motor/actuator combination was chosen, with a maximum thrust of $1000\,\mathrm{N}$ and a useful stroke of $210\,\mathrm{mm}$. The shape of the pickup bowl was made by creating a $4\,\mathrm{mm}$ offset from the shape of the tool's rotating backing pad. A hole with a diameter of $5.\mathrm{mm}$ was made in the bottom of the cup for the inlet of polish. all system components were made of 2011 aluminium alloy to reduce weight and facilitate their manufacture.
Replacing the exhausted cartridge turns out to be a quick and easy process. Basically, the process consists of:
\begin{itemize}
    \item unhooking the connection system between stick holder and bowl;
    \item unscrewing the stick from the bowl;
    \item screwing the new stick onto the bowl;
    \item hooking the connection system between stick holder and bowl.
\end{itemize}
The estimated time for cartridge replacement is about 20-30 seconds. In case it is necessary to change the color of the polishing cream, it is necessary to replace both the cartridge and the cup (several cups are provided, one for each color).
As already anticipated, the polish pickup process also requires homogenisation of the polish distribution on the robot tool. This process is performed on an additional bowl with a larger radius of curvature, where the robot rubs and tilts the tool in all directions. The two steps are illustrated in Figure \ref{fig:subfig prelievo e omog}.

\subsection{Force Control}
The control of the force applied by the robot on the external environment can be passive or active. In any case, the robot must be compliant in the direction perpendicular to the surface to ensure a constant polishing force, achieving uniform polishing across the entire shoe. In the present application, active force control based on feedback from the load cell mounted on the robot flange was chosen. After several attempts, it was seen that the correct force to be exerted is $5\,\mathrm{N}$.
The Universal Robot controller offers some special tools to control the contact force during motion: a constant force, for example, can be applied along the direction of the tool while it is moving orthogonal to that direction. However, the direction of the tool is not updated continuously, but remains fixed, equal to the direction the tool took at the time the force control was first activated during the motion. To overcome this problem, a thread running in parallel with the main program and synchronised at $500\,\mathrm{Hz}$ was implemented to handle the force control: in the thread the direction of the force to be applied is continuously updated by making it coincide with that of the tool (which is perpendicular to the surface to be polished). However, the force control tool imposes other limitations that restrict the robot's workspace, i.e. it cannot work in regions where the dexterity  (defined as $det(\mathbf{J}^T \mathbf{J})$) is lower than $10^{-3}$. In that regions, in fact, the robot approaches singular configurations and is unable to apply the required forces in some directions. When the shoe is positioned as shown in Figure \ref{fig:banco reale}, for example, the robot is unable to reach in force mode the heel area and the back side areas since it is near to wrist and head singularities. To overcome this problem a rotary table was introduced to rotate the shoe when its back side has to be worked; as a result, the robot works with high dexterity at all the points of the trajectory.

\begin{figure*}[h!]
\centering\includegraphics[width=\textwidth]{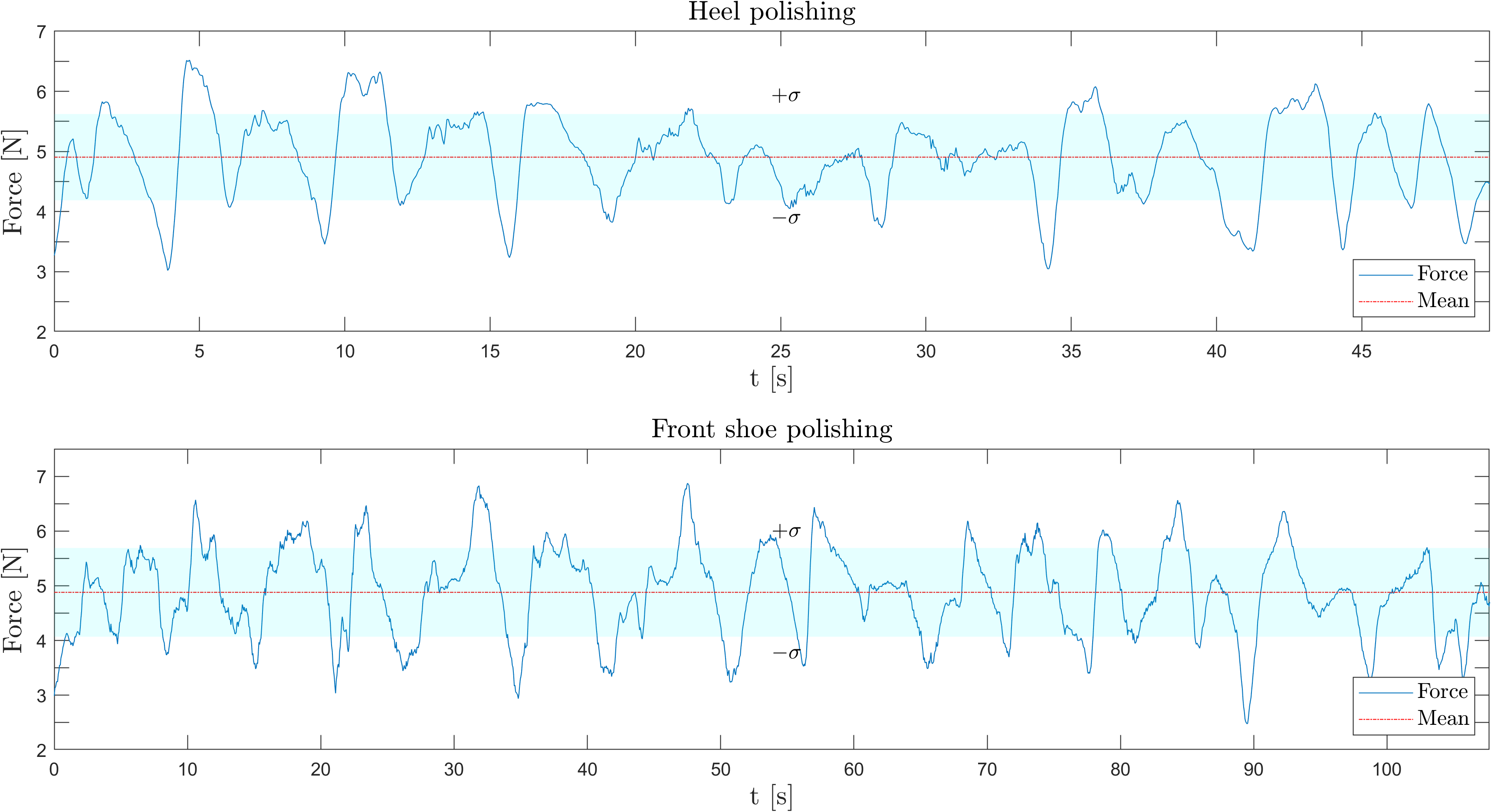}
\caption{Plot of the tool contact force versus time.}
\label{fig:grafico forze}
\end{figure*}

Figure \ref{fig:grafico forze} plots the force along the tool axis (perpendicular to the shoe surface) measured by the load cell. Specifically, data related to the front and the heel of the shoe are presented. Oscillations around the desired value of $5\,\mathrm{N}$ can be observed at very low frequency (less than $1\,\mathrm{Hz}$). Nevertheless, the mean value is very close to the target value.

\begin{figure}[h!]
\centering\includegraphics[width=7cm]{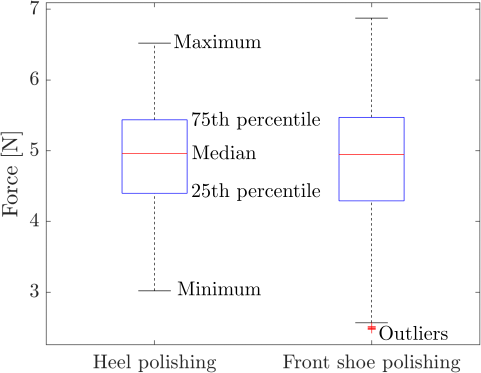}
\caption{Boxplot of tool contact force.}
\label{fig:boxplot forze}
\end{figure}

Table \ref{tab:statistica forze} and Figure \ref{fig:boxplot forze} detail all the statistical information of the force recorded in the two trajectories.

\begin{table*}[h!]
\caption{Statistical descriptors of tool contact force}
\label{tab:statistica forze}
\centering{%
\begin{tabular}{l c c c c c c}
\toprule
 & Average [N] & Median [N] & $\sigma$ [N] & $Q_1$ [N] & $Q_3$ [N] & Number Outliers\\
\midrule
Heel & 4.90 & 4.96 & 0.70 & 4.39 & 5.43 & 0 \\
Front & 4.88 & 4.94 & 0.80 & 4.29 & 5.47 & 5 \\
\bottomrule
\end{tabular}
}%
\end{table*}

\subsection{Polish Trajectories}

The trajectories executed by the tool during the polishing process have a major influence on the quality of the final result. For this purpose, the shoe was divided into sectors and trajectories were designed for each of them using the CAM software Siemens NX. Specifically, the shoe was divided in the following sectors: toe, heel, the two side parts and the upper. It is important that each area is polished with a appropriate number of trajectories (4 for the toe, 2 for the heel, 2 for each side), but it is important to find a right trade-off between the number of polishing trajectories (more trajectories would result in better quality) and the working time.
These trajectories must be different so as not to leave indelible streaks and polish marks that permanently compromise the quality of the polish, even to the point of discarding the shoe. Of course, special attention should be paid to the toe area, which is what the end customer appreciates most. The trajectories for each sector were designed following different patterns, such as zigzag or spiral. In addition, to ensure differentiation, the pitch followed in the zigzag or spiral pattern was varied between $5,\mathrm{mm}$ and $8,\mathrm{mm}$. Also starting and end points of each trajectory should be varied. A CAD model of the shoe is needed to generate accurate trajectories that can meet some critical requirements, such as:

\begin{itemize}
    \item the tool must maintain a perpendicular orientation to the shoe surface at all times, ensuring a consistent force across the entire worked area.;
    \item the pitch between adjacent lines of a trajectory must be constant and evaluated in relation to the size of the polishing pad;
    \item in the parts near the sole, the pad should come as close as possible but not contact the sole to prevent damage.
\end{itemize}

For all these reasons, the assumption that trajectories can be taught to the robot directly by the operator through manual tool guidance and path registration was considered, but then discarded. In fact, the operator, although equipping the robot with a special handle to guide, fails to ensure for each trajectory all the requirements necessary to achieve optimal polishing. 

\begin{figure}
\centering
\begin{subfigure}{.23\columnwidth}
  \includegraphics[width=\columnwidth]{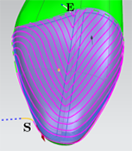}  
  \caption{}
\end{subfigure}
\begin{subfigure}{.23\columnwidth}
  \includegraphics[width=\columnwidth]{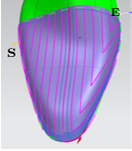}  
  \caption{}
\end{subfigure}
\begin{subfigure}{.23\columnwidth}
  \includegraphics[width=\columnwidth]{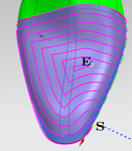}  
  \caption{}
\end{subfigure}
\begin{subfigure}{.23\columnwidth}
  \includegraphics[width=\columnwidth]{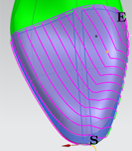}  
  \caption{}
\end{subfigure}
\caption{Different type of polishing trajectories for the toe.}
\label{fig: subfig traiettorie}
\end{figure}

Figure \ref{fig: subfig traiettorie} shows 4 types of trajectories generated for the toe. It can be observed that the start (S) and end (E) points are different in all cases, such as the path pattern. Finally, the trajectories generated for the entire shoe are 6 for the toe and upper area, 2 for each side area, and 3 for the heel. 

\begin{figure*}[htb]
\centering
\includegraphics[width=\textwidth]{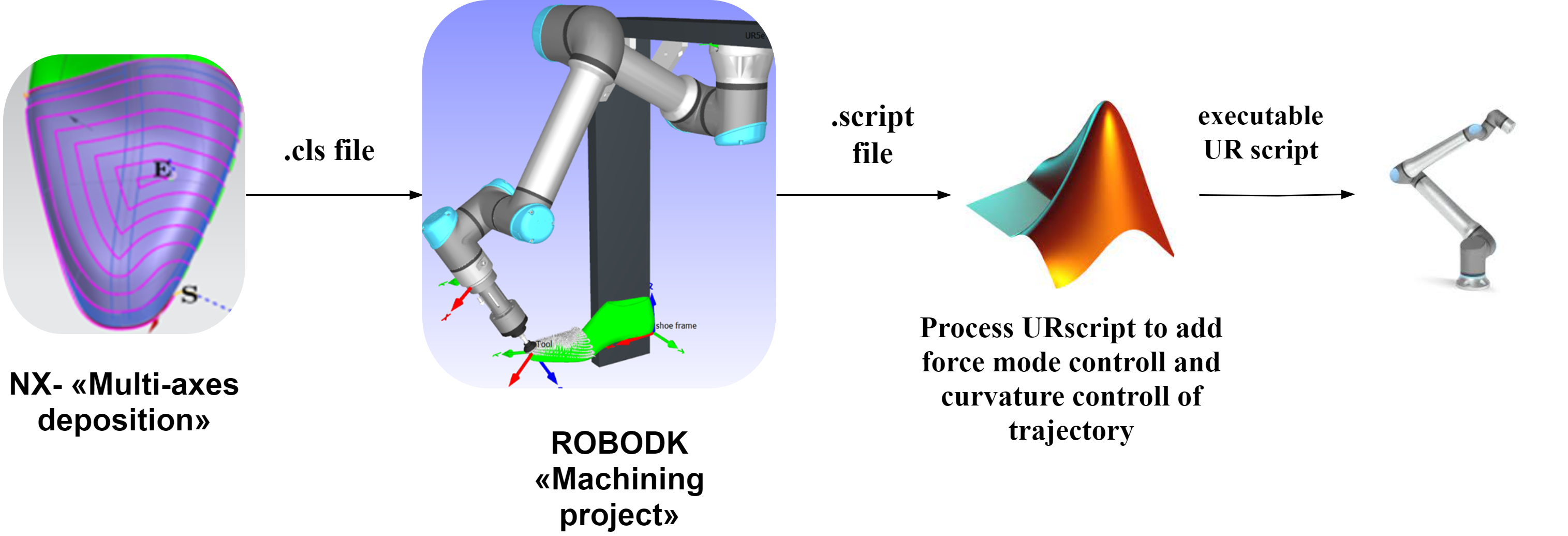}
\caption{Process to obtain the executable UR script.}
\label{fig:diag processo script}
\end{figure*}

Figure \ref{fig:diag processo script} shows the implemented process that leads from the concept of trajectories to their implementation. In particular, the process starts from the NX production software where the "Multi axes deposition" function is used to define the trajectory from the CAD of the shoe, choosing different paths, different guide lines and different start and end points. A .cls file is then exported where the trajectory points and orientations are stored. The .cls file is imported to ROBODK where the "Machining Project" module is used to generate the robot program. ROBODK is able to generate the executable .script file for UR robots. The file is post-processed through MatLab to add commands related to force control and trajectory curvature which is used to manage the motion of the tool center point. The time required for the entire trajectory generation process, from design to the generation of final UR executable file, depends on the number of trajectories and the operator's skill level. In the worst-case scenario, with 10 trajectories for the toe and vamp, 2 for each lateral side, and 4 for the heel, the process can take up to 1 hour. On the other hand, once these trajectories are created, their generation for the opposite shoe is virtually instantaneous.

\begin{figure}[htbp]
\centering\includegraphics[width=8cm]{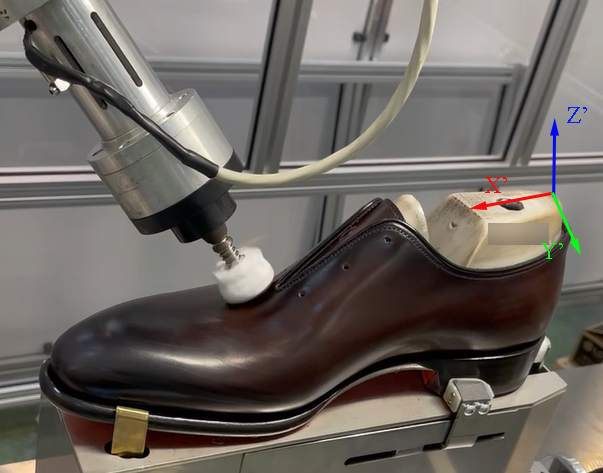}
\caption{Reference system $\mathrm{Ox'y'z'}$ measured by the robot.}
\label{fig:ref syst su forma}
\end{figure}

\begin{figure}[h]
\centering
\begin{subfigure}{.6\columnwidth}
  \includegraphics[width=\columnwidth]{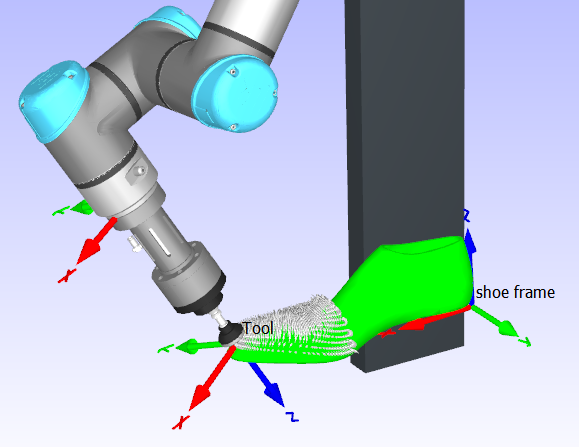}  
  \caption{ROBODK reference system.}
  \label{subfig: ROBODK ref system}
\end{subfigure}
\hfill
\begin{subfigure}{.35\columnwidth}
  \includegraphics[width=\columnwidth]{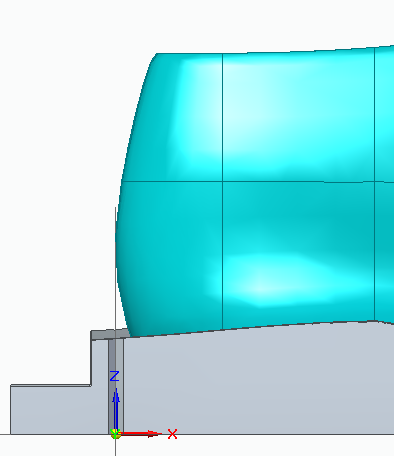}  
  \caption{Calibration support reference system.}
  \label{subfig: supporto scarpa}
\end{subfigure}
\caption{Shoe reference frame.}
\label{fig: subfig robodk ref system}
\end{figure}

To assess the reachability of all points of the shoe with the adopted setup, the trajectories for the largest and smallest size present in the collection were plotted from the CAD of the respective last. The system is able to polish all sizes present in the collection, for both the right and left shoe, and all shoe models based on the same last. The trajectories are generated at first for the right shoe. Then, due to the specularity, to obtain the trajectory of the left shoe it is sufficient to invert the sign of the $\mathrm{Y}$ and $\mathrm{Ry}$ values stored in the .cls file.\\
In order to generate the trajectory, it is necessary to know where the reference frame of the shoe is located in the real workstation so as to indicate it in ROBODK and generate the points of the trajectory referenced with respect to that frame.
The position of the shoe reference frame with respect to the robot base is needed to transform the coordinates of the trajectories from local to robot frame. To this aim, once the shoe is mounted on the appropriate stand, the position and orientation of the reference system $\mathrm{Ox'y'z'}$ (Figure \ref{fig:ref syst su forma}) is acquired by contact with the robot. Known the position of the reference system $\mathrm{Ox'y'z'}$ the pose of the reference system $\mathrm{Oxyz}$ (Figure \ref{subfig: ROBODK ref system}) positioned under the heel and not directly measurable, is obtained using a custom calibration equipment: a support was constructed that perfectly replicates the shape of the heel of the last has been prototyped; thanks to this support it is easy to measure where the system shown in Figure \ref{subfig: supporto scarpa}, coincident with $\mathrm{Oxyz}$, is positioned. Thus, once the pose of the $\mathrm{Ox'y'z'}$ system is acquired, the relative pose between the two systems is known. It is therefore simply necessary to acquire for each size only the pose of the $\mathrm{Ox'y'z'}$ system on the bench to consequently know the pose of $\mathrm{Oxyz}$, with respect to which the points of the trajectory are given.

\subsection{Speed of Movements}

The speed of movement of the TCP on the leather must be as constant as possible to ensure an even polish over the entire surface of the shoe. The optimal feed speed of the polishing tool is determined by a trade-off: too high a speed prevents the polish from penetrating the pores of the leather properly, while too slow a speed causes too long a processing time, which must not exceed the time taken for manual polishing. A value of $v_0=0.03\,\mathrm{m/s}$, determined by trial and error, guarantees the best compromise between quality and processing time. The Universal Robot control offers two types of commands to control the motion over a predefined Cartesian path, i.e. \textit{moveL} for straight trajectories and \textit{moveP} for non-straight trajectories with constant velocity of the TCP along the path (e.g. for welding). Thus, \textit{moveP} revealed the right command to be used in the polishing operation. Since the shape of the shoe surface assumes rather different radii of curvature along the various parts of a polishing trajectory, maintaining a constant speed along the path can lead to two cases:
\begin{itemize}
    \item If the speed is too high, at points with a very small radius of curvature (such as the toe of the shoe) the robot goes into a safety stop because, in order to ensure a constant linear speed, the first three joint speeds exceed the maximum permitted value.
    \item If the speed is too low, the robot is able to perform the motion in sections with a small radius of curvature, but in sections with a large radius of curvature it is too slow, greatly increasing the cycle time..
\end{itemize}

Therefore, to solve this problem an adaptive tool speed control is designed based on the curvature of the trajectory that the robot is executing at that instant. The radius $r$ of the oscillating circle passing through the two points just executed by the robot and the next one about to be executed is calculated. Then, the speed assumed by TCP is calculated as:

\begin{equation}
v_{TCP}=\frac{v_0}{\lambda}
\end{equation}

where $v_0$ is the reference speed assumed in linear sections of the trajectory and being 

\begin{equation}
\left\{
\begin{array}{l l}
\lambda=red_{value}+k\left(r_{lim}-r\right) & r<r_{lim}\\
\lambda=1 & r\geq r_{lim}
\end{array}
\right.
\end{equation}

Thus, the speed is reduced only if the radius of curvature is minor than a threshold, which is set in the tests at $r_{lim}=0.06\,\mathrm{m}$, whereas $red_{value}$ and $k$ are two parameters that can be set according to the desired speed reduction. When possible, setting $red_{value}=1$ would lead to a smooth transition. A similar scaling law is used also for the acceleration of the TCP. In this way, it is possible to work at constant velocity $v_0$ (set at $v_0=0.03\,\mathrm{m/s}$) the parts of the shoe with reduced curvature, reducing the speed when areas with high curvature are worked in order to avoid safety stops that would lead to unrecoverable damage to the shoe. 

\section{Experimental Tests} \label{sec:exp test}

This section reports the results of various polishing tests performed on different shoes using the methodology described above. In particular, the first part describes the setting operations that the operator must perform before starting the operation, while the second part describes the results obtained.

\subsection{Process Implementation}
In the setup phase the operator must carry out a machine setting procedure. First, the shoe must be placed in the last and fixed to the stand: a knob allows to adjust the length of the clamping system to the size of the shoe, then pneumatic hooks are activate to ensure the clamp. When necessary the operator should also change the colour of the polish, the polish pickup bowl, the rubbing pad and  the rotating backing pad. 
Once the shoe is attached to the holder, the operator selects via the HMI the model and shoe number and sets whether the it is right or left. The system retrieves the basic polishing recipe for that shoe from the database, namely:

\begin{itemize}
    \item the type and number of trajectories to be performed for each area;
    \item when to perform the polish picking;
    \item the force to be exerted by the robot and the speed of polishing.
\end{itemize}

The operator's experience remains essential in this process, in order to decide whether to maintain the basic polishing recipe already validated on other shoes or whether, for that particular type of leather, to modify the processing parameters. The operator can programme to have certain areas of the shoe re-polished if the quality obtained does not meet requirements or intervene manually and correct in real time, simultaneously with the robotic operation. In this way, the skills and added value provided by the craftsmen are preserved and enhanced. While the robot completes the first polishing cycle, the operator can concentrate on the second coat of finishing polishing on shoes already worked by the robot; in fact, the final polishing, less tiring for the operator but more critical in terms of quality, is carried out completely by hand until the desired final aesthetic appearance is obtained.

\subsection{Results}

Several tests were carried out on various shoes to find out which basic recipe should be used to achieve an optimal polish. The evaluation criteria used were:
\begin{itemize}
    \item  \textbf{visual check} - evaluation of the aesthetic appearance through an empirical check of the gloss level, colour homogeneity and possible presence of stains, streaks and colour accumulation points;
    \item \textbf{touch control} - assess whether the shoe is oily or dry to the touch; dry is the optimal result;
    \item \textbf{cracking test} - control of the possible release of product residue as a result of the mechanical twisting of the shoe; less product released equals better results;
    \item \textbf{aging test} - evaluation of the ability to revive the shoe's lustre after a certain period of time (72 hours).
\end{itemize}

In addition, the ease of performing the second polishing coat by the operator (time and effort required) was assessed.
The company's chemical laboratory assigned a score for each of the above parameters according to the following Likert scale: 1 - very dissatisfied, 2 - dissatisfied, 3 - slightly satisfied, 4 - satisfied, 5 - very satisfied.
Due to confidentiality agreements, some specific values used to assess the quality of the process cannot be disclosed. Tables \ref{tab risultati 1} and \ref{tab risultati 2} report the main process parameters for the different tests and qualitatively summarises the satisfaction of the evaluation criteria.

\begin{table}[]
    \centering
    \begin{tabular}{c|c|c}
    \toprule
         & \textbf{Shoe 1} & \textbf{Shoe 2} \\
    \midrule
    \multirow{3}{*}{Number of Trajectories} & \multirow{1}{*}{Toe and Vamp: 10} & \multirow{1}{*}{Toe and Vamp: 6}\\
    & \multirow{1}{*}{Lateral (each side): 2} & \multirow{1}{*}{Lateral (each side): 2}\\
     & \multirow{1}{*}{Heel: 4} & \multirow{1}{*}{Heel: 2}\\
     \midrule
    Number of Polish Picking & 9 & 8 \\
    \midrule
    Cycle Time & 1 (40 minutes) & 3 (28 minutes) \\
    \midrule
    Visual Check & 5 & 5 \\
    \midrule
    Touch Check & 2 & 4 \\
    \midrule
    Cracking Test & 5 & 5 \\
    \midrule
    Ageing Test & 5 & 5 \\
    \midrule
    Second Hand Polishing & 5 & 5 \\
    \bottomrule
    \end{tabular}
    \caption{Process parameters and quantitative results of the different tests 1 and  2.}
    \label{tab risultati 1}
\end{table}

\begin{figure*}[h!]
\centering
\begin{subfigure}{.48\textwidth}
  \includegraphics[width=\textwidth]{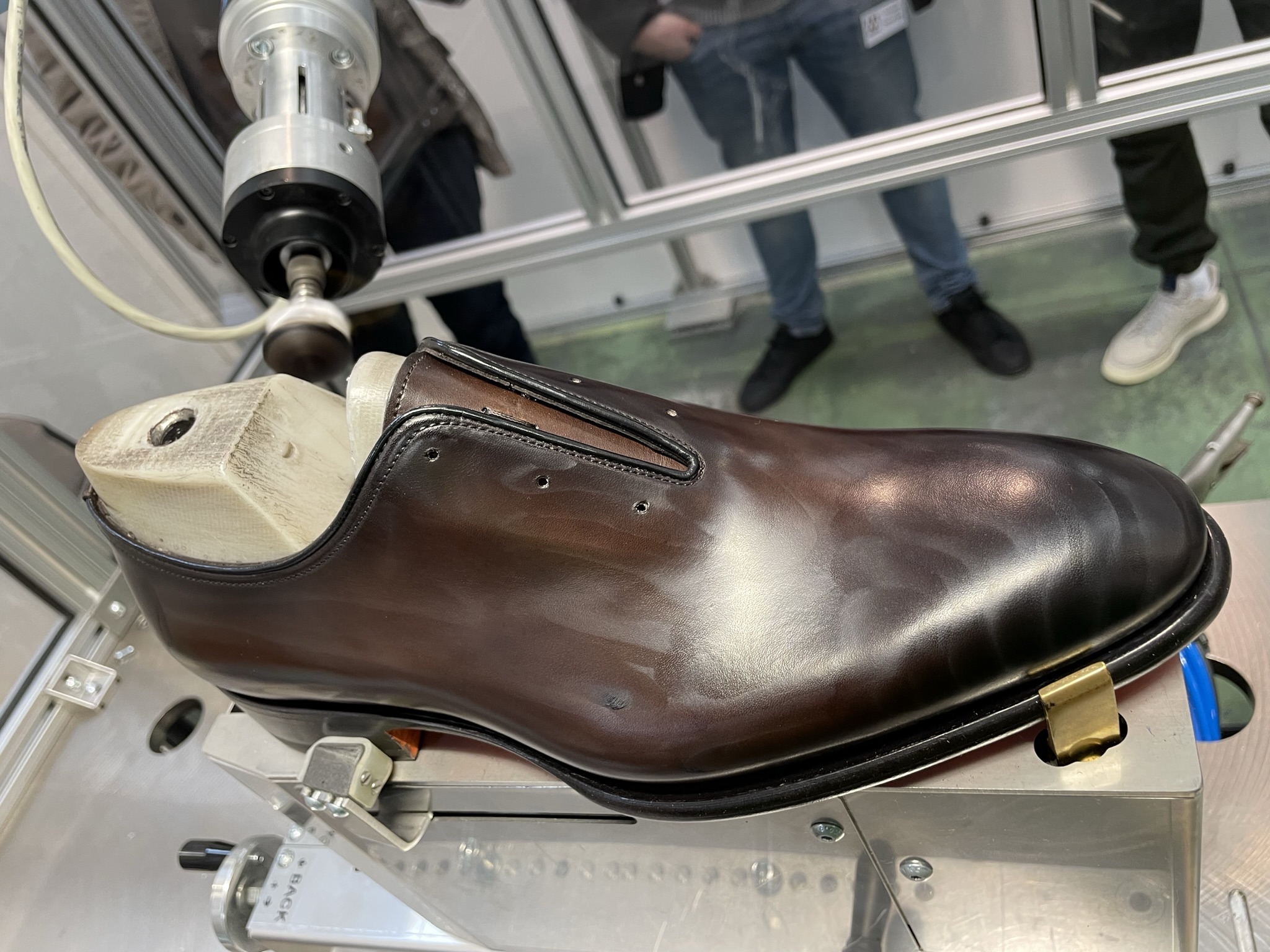}  
  \caption{Shoe 1}
\end{subfigure}
\begin{subfigure}{.48\textwidth}
  \includegraphics[width=\textwidth]{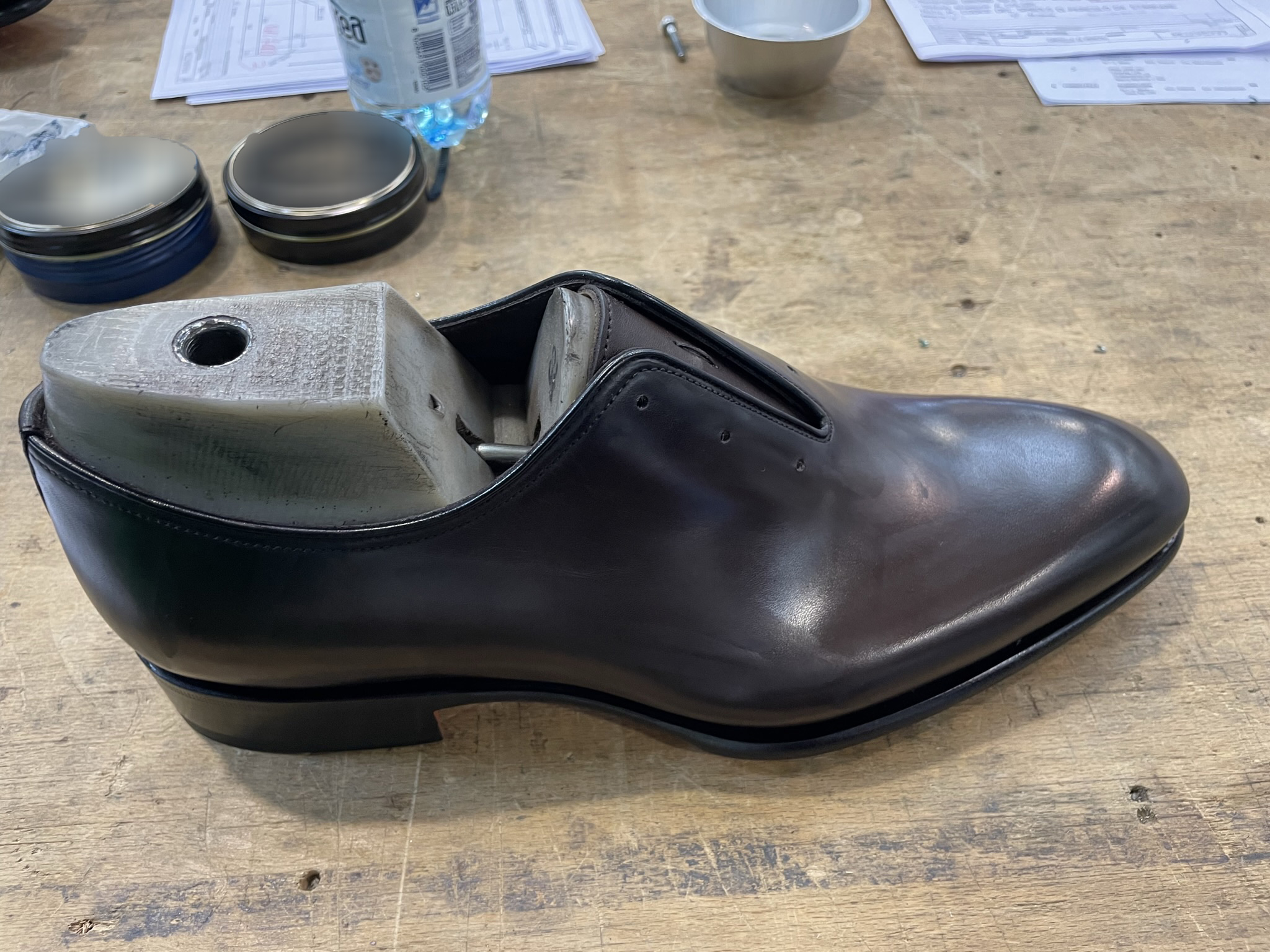}  
  \caption{Shoe 2}
\end{subfigure}

\begin{subfigure}{.48\textwidth}
  \includegraphics[width=\textwidth]{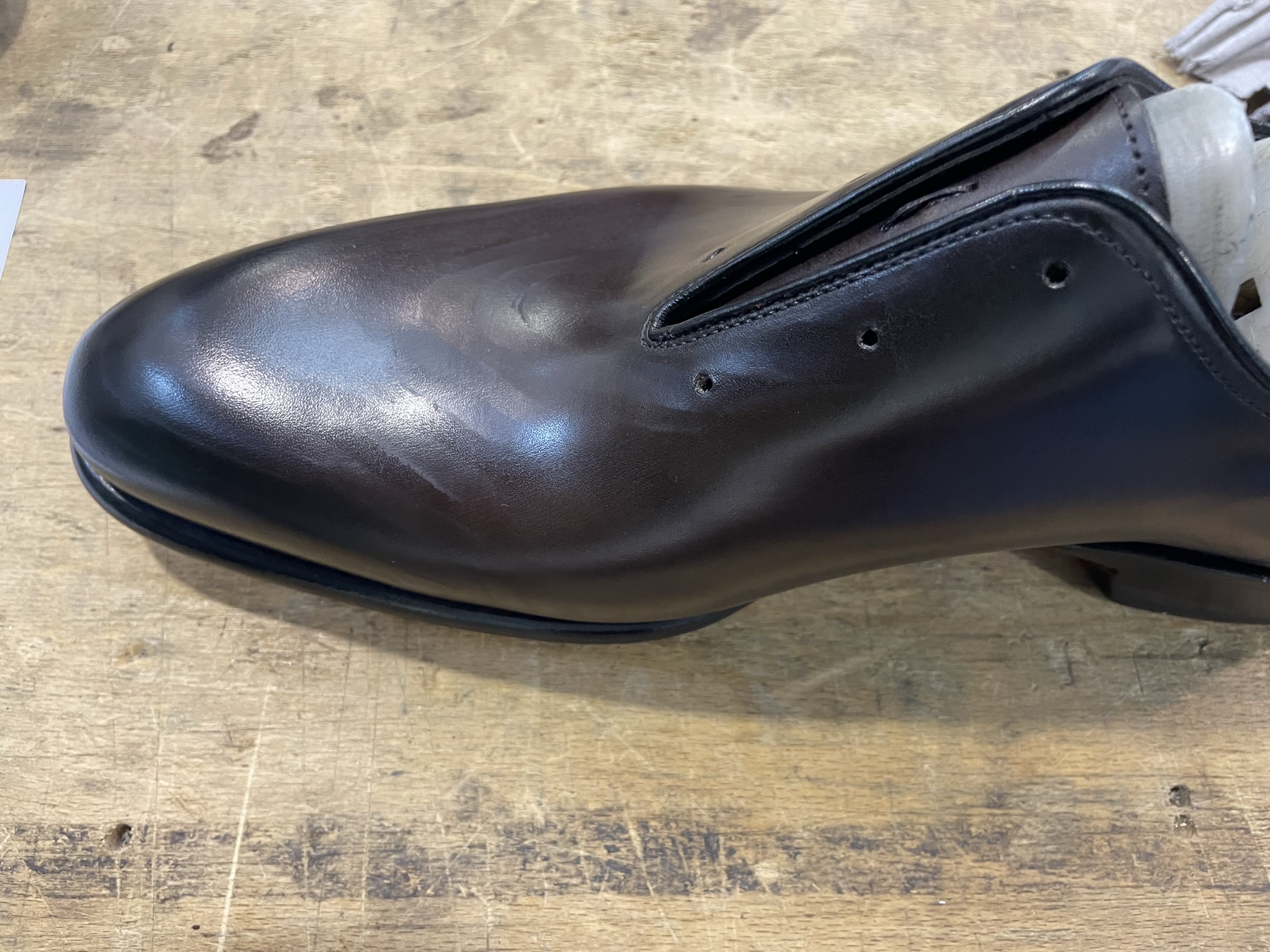}
  \caption{Shoe 3}
\end{subfigure}
\begin{subfigure}{.48\textwidth}
  \includegraphics[width=\textwidth]{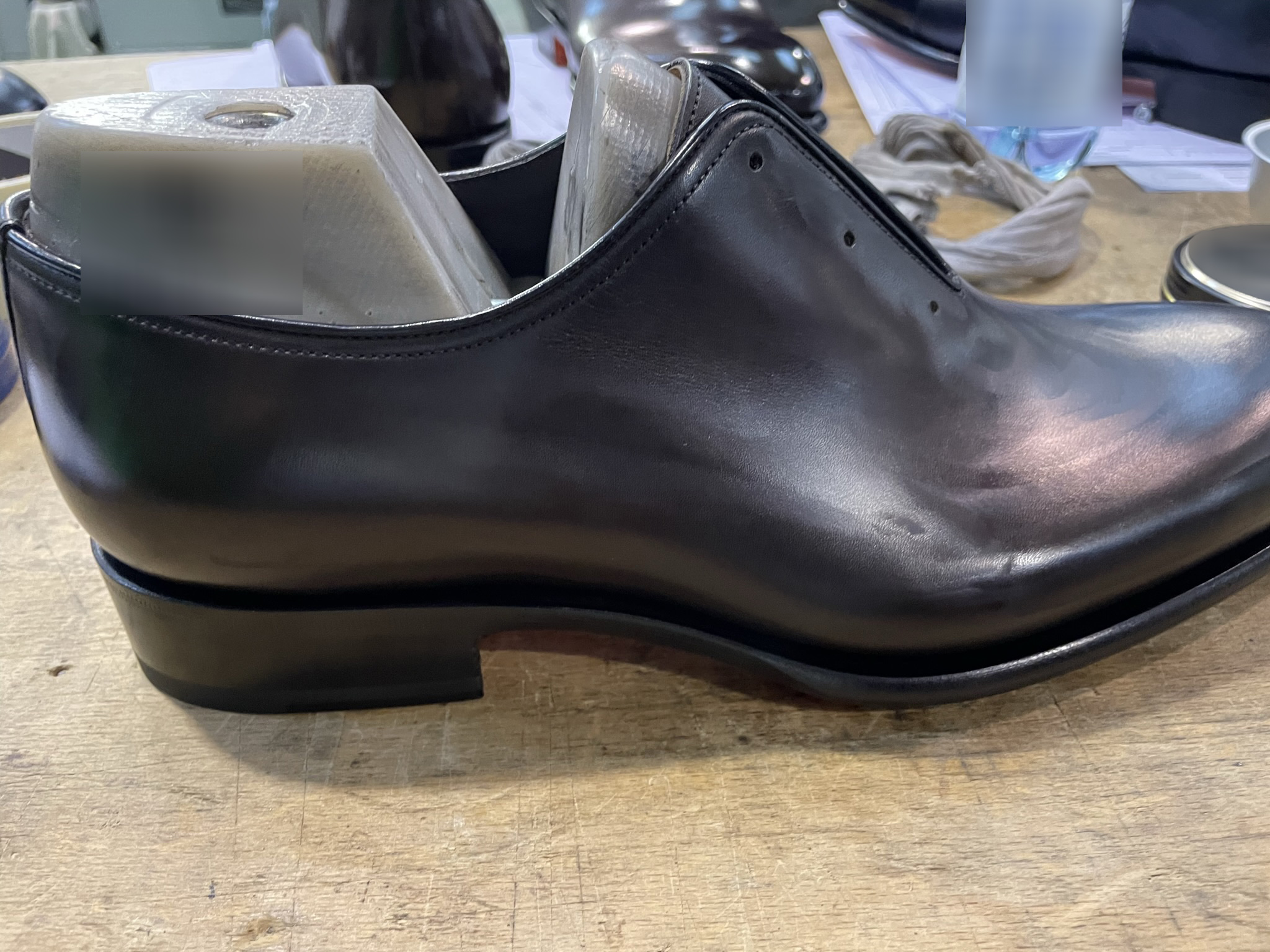}
  \caption{Shoe 4}
\end{subfigure}
\caption{Tested shoes after first phase of polishing performed by the robot.}
\label{fig: subfig scarpe 1 mano}
\end{figure*}

\begin{table}[]
    \centering
    \begin{tabular}{c|c|c}
    \toprule
         & \textbf{Shoe 3} & \textbf{Shoe 4}\\
    \midrule
    \multirow{3}{*}{Number of Trajectories} & \multirow{1}{*}{Toe and Vamp: 4} & \multirow{1}{*}{Toe and Vamp: 4}\\
   & \multirow{1}{*}{Lateral (each side): 2} & \multirow{1}{*}{Lateral (each side): 2}\\
     & \multirow{1}{*}{Heel: 2} & \multirow{1}{*}{Heel: 2}\\
     \midrule
    Number of Polish Picking & 7 & 5\\
    \midrule
    Cycle Time &  4 (20 minutes) & 5 (18 minutes)\\
    \midrule
    Visual Check &  4 & 3\\
    \midrule
    Touch Check &  5 & 2\\
    \midrule
    Cracking Test &  3 & 2\\
    \midrule
    Ageing Test &  3 & 3\\
    \midrule
    Second Hand Polishing & 3 & 1\\
    \bottomrule
    \end{tabular}
    \caption{Process parameters and quantitative results of the different tests 3 and  4.}
    \label{tab risultati 2}
\end{table}
The second test (Shoe 2) meets all quality requirements; it is slightly oily to the touch after the first polishing step, but this factor helps the operator in the second polishing step.
Pictures of the four tested shoes after the first coat of polish are reported for completeness in Figure \ref{fig: subfig scarpe 1 mano}.

\section{Discussion} \label{sec:discussion}

Tests demonstrate the feasibility of automating the polishing process for luxury shoes. In particular, shoe 2, although taking 10 minutes longer than the time taken by the operator, fully meets all the aesthetic and physical criteria of polishing. The result obtained is even better than that obtainable manually, as the backing pad, rotating at high speed, heats the leather and makes the polish penetrate deeper, with obvious benefits in the crack test.\\
Shoe 3 requires less polishing time than the time taken by the operator by hand (1-2 minutes less); however, the quality is slightly lower than shoe 2. This is due to the reduced number of polishing trajectories performed on it. After the final polishing, the shoe still passes all the aesthetic checks, but the final operation requires more effort to compensate for the less work carried out in the first phase. Thus, it is a company's choice whether to aim for better quality in the first phase, but increase the cycle time, or reduce the cycle time requiring more manual processing in the second phase.\\
Shoe 1 was over-processed with too many trajectories and too much polish input; the cycle time is very high and the shoe feels oily to the touch. Therefore, less processing was chosen for the other tests.\\
In shoe 4, a further reduction in cycle time was achieved by decreasing the polishing picking, however, the results did not meet the desired quality even after the second polishing cycle.\\
In Figure \ref{fig: subfig scarpe 1 mano} some streaks of polish, present at the end of the first phase, are visible. These streaks can be easily removed with the second polishing phase, which gives the shoe a perfect homogeneity. This is only possible if the polishing trajectories in the first phase are heterogeneous, otherwise there is a risk that the streaks may be permanent.\\
Tests have shown that a key element for polishing success is the wear of the cotton tissue mounted on the rotating pad: to meet standards, the tool must be replaced after 6 shoes. This can be done thanks to a quick-release mechanism that allows for quick and easy replacement.\\
The same process parameters were used for another shoe model and colour, but mounted on the same last, achieving good polishing results, demonstrating the transferability of the operation to different shoe models. 

This robotic polishing process is worthwhile if the number of shoes to be polished justifies the initial time investment of about 1 hour to prepare all the trajectories. 
Given a time savings of about 2 minutes per shoe, the process becomes efficient if the number of shoes to be polished using the same last (including both right and left shoes) exceeds 30 single shoes.
Since producing a fully customized pair for a single customer is rare for the project partner footwear company, the strategy has been deemed feasible for their needs.
The primary benefit to the company is not just a slight improvement in productivity but, more importantly, the enhanced well-being of the operators. By relieving them from stressful and physically demanding tasks, the process helps reduce the risk of work-related injuries and illnesses. While the robot performs its task, the operator can carry out the second-hand polish, which is less stressful and requires less finger pressure on the leather. This process takes about 8 minutes. During the remaining time, the operator can manage the robot station, check the quality of the robot's work by entering the cell, and manually fix any polish imperfections with a cotton cloth.

When comparing the proposed approach with that presented by \citet{borrell2023cooperative}, several key differences emerge, along with the introduction of innovative aspects. First of all in \citet{borrell2023cooperative} the polishing trajectories are manually  recorded, moving the TCP of the robot on the shoe surface and storing the points of the path. The recorded trajectory is then imported into CAM software to verify its feasibility and correct any imperfections, such as missing points or TCP orientation. In contrast, the proposed approach generates the trajectories directly within the CAM software, leading to several positive outcomes, which can be summarized as follows:
\begin{itemize}
\item utilising the last as a geometric model offers greater flexibility, allowing the same trajectories to be applied across various footwear models that use the same last; the control system compensates for differences between the last and the shoe upper by processing data from the force/torque sensor;
\item the software automatically generates the path based on operator inputs, such as surface, direction, and line spacing, making the process highly efficient and user-friendly; operators can also modify trajectories to meet specific requirements;
\item the path for the opposite shoe is automatically generated through symmetry;
\item path generation for different shoe sizes is semi-automated, starting from the trajectory of the sample size;
\item time and errors associated with learning the path are significantly reduced; skilled operators with expertise in designing trajectories using the robot's TCP are no longer required; after some training, a technician can easily operate the software and generate trajectories efficiently.
\end{itemize}
Additionally, the proposed approach incorporates several innovative features, including:
\begin{itemize}
    \item precise control of the amount of polish applied;
\item adjusted speeds of the polishing tool in areas of the shoe with high curvature;
\item an automatic shoe locking system, which provides precise knowledge of the shoe's reference frame after an initial calibration, eliminating the need for repetitive measurements by vision systems.
\end{itemize}
All these factors contribute to achieving a polishing quality comparable to hand polishing. However, a direct comparison with \citet{borrell2023cooperative}  in terms of polishing quality is challenging due to potential differences in baseline aspects, such as the type of polish used, the leather type, and certain implementation parameters that were not disclosed.

\section{Conclusions} \label{sec:conclusion}


This paper presented a collaborative robotic cell for the polishing of luxury leather shoes. The solution was realised in response to the need of companies in the footwear sector to automate processes currently carried out entirely by hand, in order to improve product quality and productivity, while enhancing the well-being of the operator.\\
Shoe companies, like many other manufacturing companies, need to move towards human-centred production in order to remain competitive and at the same time guarantee high quality standards. 
The proposed technological solution goes in this direction: the operator is relieved of tiring and repetitive work, assuming the role of a supervisor who, based on his experience, can modify the process parameters to meet the required standards in real-time and fix any polishing imperfection by getting inside the cell and manually work simultaneously to the robotic operation. At the same time, the operator can devote more time to the final finish, where manual dexterity and craftsmanship are the real added value of the process.\\
The preliminary analysis on the key factors that influence the process, such as the force exerted by the robot, the speed of movement, the types of trajectories and the quantity of polish used, was necessary in order to develop a process compliant with the required quality standards.\\
After the description of the design of the robotic cell and tools, the methodology used to set the process parameters was illustrated. The results obtained equals manual polishing in terms of quality. As regards the cycle time, a reasonable but not optimal result was obtained (about 30 minutes), higher than the time taken by an operator (about 20-23 minutes) if optimal quality is to be achieved. However, the longer processing time and a different way of applying the polish allow for optimal penetration of the cream, which facilitates subsequent manual finishing.
Alternatively, to reduce manufacturing time in the initial phase, the solution proposed for Shoe Test 3 could be implemented. This would ensure quality standards are met, although the second polishing would be slightly more strenuous for the operator.  Finally, the possibility of working with both different sizes and different shoe models has been demonstrated.\\
In the authors' opinion, this work has a significant impact on the development of footwear companies. Today, the well-being of operators is a critical concern, and polishing is one of the most stressful and physically demanding tasks. Companies face serious challenges in finding workers willing to perform this operation due to the harsh working conditions, as well as a high incidence of injuries and musculoskeletal disorders that affect productivity and staff management. The proposed solutions address these issues by automating the most tedious phase of polishing with a robot, maintaining the same level of quality while elevating the operator to a supervisory role with less stressful activities.
The main limitation of the proposed solution is that the trajectories must be obtained from CAD for each shoe shape and size; this involves a considerable amount of off-line work. A possible strategy to limit this problem could be the implementation of some geometric scaling rule to automatically generate the various sizes of a model. Currently the only automatic procedure concerns the generation of the trajectories of the left shoe starting from the right one through the application of a simple mirror symmetry.\\
As a future development, a more in-depth analysis of the process parameters will be conducted, also through an extensive test campaign, to try to reduce the cycle time at least to the limit of manual operation. A possible solution to have a significant productivity increasing would be to implement two robots, in order to process the right and left shoe simultaneously, halving the overall cycle time, some preliminary simulations test have been studied, promising a big impact on the productivity. With the solution being studied, it will be possible to achieve high quality standards, with significantly less time and ensuring the well-being of the operator. An innovative pad will also be studied, less subject to wear, so that replacement operations are limited, further speeding up the process.

\section*{Funding}
No funding was received to assist with the preparation of this manuscript.
\section*{Disclosure statement.}
All authors certify that they have no affiliations with or involvement in any organization or entity with any financial interest or non-financial interest in the subject matter or materials discussed in this manuscript.
The authors consent to the publication

\section*{Data availability statement}
Data, Materials, Code related to this work will be provided upon request.

\bibliographystyle{tfcad}
\bibliography{main.bib}

\begin{thebibliography}{22}
\newcommand{\enquote}[1]{``#1''}
\providecommand{\natexlab}[1]{#1}
\providecommand{\url}[1]{\normalfont{#1}}
\providecommand{\urlprefix}{}

\bibitem[Borrell et~al.(2023)]{borrell2023cooperative}
Borrell, Jorge, Alejandra Gonz{\'a}lez, Carlos Perez-Vidal, Luis Gracia, and J~Ernesto Solanes. 2023. ``Cooperative human--robot polishing for the task of patina growing on high-quality leather shoes.'' \emph{The International Journal of Advanced Manufacturing Technology} 125 (5-6): 2467--2484.

\bibitem[Castelli et~al.(2020)]{castelli2020feasibility}
Castelli, Kevin, Ahmed Magdy~Ahmed Zaki, Yevheniy Dmytriyev, Marco Carnevale, and Hermes Giberti. 2020. ``A feasibility study of a robotic approach for the gluing process in the footwear industry.'' \emph{Robotics} 10 (1): 6.

\bibitem[Chiriatti et~al.(2022)]{chiriatti2022human}
Chiriatti, Giorgia, Marianna Ciccarelli, Matteo Forlini, Melania Franchini, Giacomo Palmieri, Alessandra Papetti, and Michele Germani. 2022. ``Human-centered design of a collaborative robotic system for the shoe-polishing process.'' \emph{Machines} 10 (11): 1082.

\bibitem[Choi, Hwang, and You(2008)]{choi2008development}
Choi, Hyeung-Sik, Gyu-Deuk Hwang, and Sam-Sang You. 2008. ``Development of a new buffing robot manipulator for shoes.'' \emph{Robotica} 26 (1): 55--62.

\bibitem[Cocuzza, Fornasiero, and Debei(2013)]{cocuzza2013novel}
Cocuzza, Silvio, Rosanna Fornasiero, and Stefano Debei. 2013. ``Novel automated production system for the footwear industry.'' In \emph{Advances in Production Management Systems. Competitive Manufacturing for Innovative Products and Services: IFIP WG 5.7 International Conference, APMS 2012, Rhodes, Greece, September 24-26, 2012, Revised Selected Papers, Part I}, 542--549. Springer.

\bibitem[Gracia et~al.(2017)]{gracia2017robotic}
Gracia, Luis, Carlos Perez-Vidal, Dennis Mronga, Jose-Manuel de~Paco, Jose-Maria Azorin, and Jose de~Gea. 2017. ``Robotic manipulation for the shoe-packaging process.'' \emph{The International Journal of Advanced Manufacturing Technology} 92: 1053--1067.

\bibitem[Hu et~al.(2007)]{hu2007automatic}
Hu, Zhongxu, Chris Marshall, Robert Bicker, and Paul Taylor. 2007. ``Automatic surface roughing with 3D machine vision and cooperative robot control.'' \emph{Robotics and Autonomous Systems} 55 (7): 552--560.

\bibitem[Jatta et~al.(2004)]{jatta2004roughing}
Jatta, Francesco, Lorenzo Zanoni, Irene Fassi, and S~Negri. 2004. ``A roughing/cementing robotic cell for custom made shoe manufacture.'' \emph{International Journal of Computer Integrated Manufacturing} 17 (7): 645--652.

\bibitem[Kim et~al.(2022)]{kim2022robot}
Kim, Min-Gyu, Juhyun Kim, Seong~Youb Chung, Maolin Jin, and Myun~Joong Hwang. 2022. ``Robot-based automation for upper and sole manufacturing in shoe production.'' \emph{Machines} 10 (4): 255.

\bibitem[Lee, Kao, and Wang(2018)]{lee2018implementation}
Lee, Chun-Yi, Tsai-Ling Kao, and Ko-Shyang Wang. 2018. ``Implementation of a robotic arm with 3D vision for shoes glue spraying system.'' In \emph{Proceedings of the 2018 2nd International Conference on Computer Science and Artificial Intelligence}, 562--565.

\bibitem[Lu et~al.(2022)]{lu2022outlook}
Lu, Yuqian, Hao Zheng, Saahil Chand, Wanqing Xia, Zengkun Liu, Xun Xu, Lihui Wang, Zhaojun Qin, and Jinsong Bao. 2022. ``Outlook on human-centric manufacturing towards Industry 5.0.'' \emph{Journal of Manufacturing Systems} 62: 612--627.

\bibitem[Maurtua, Ibarguren, and Tellaeche(2012)]{maurtua2012robotic}
Maurtua, I{\~n}aki, Aitor Ibarguren, and Alberto Tellaeche. 2012. ``Robotic solutions for footwear industry.'' In \emph{Proceedings of 2012 IEEE 17th International Conference on Emerging Technologies \& Factory Automation (ETFA 2012)}, 1--4. IEEE.

\bibitem[M{\'e}ndez et~al.(2020)]{mendez2020robotic}
M{\'e}ndez, Jorge~Borrell, Carlos Perez-Vidal, Jos{\'e} Vicente~Segura Heras, and Juan~Jos{\'e} P{\'e}rez-Hern{\'a}ndez. 2020. ``Robotic pick-and-place time optimization: Application to footwear production.'' \emph{IEEE Access} 8: 209428--209440.

\bibitem[Nemec and {\v{Z}}lajpah(2006)]{nemec2006automation}
Nemec, Bojan, and Leon {\v{Z}}lajpah. 2006. ``Automation in shoe assembly.'' In \emph{Proceedings of the 15th International Workshop on Robotics in Alpe-Adria-Danube Region (RAAD 2006), Balatonf{\"u}red, Hungary}, 131--136.

\bibitem[Nemec and {\v{Z}}lajpah(2008)]{nemec2008robotic}
Nemec, Bojan, and Leon {\v{Z}}lajpah. 2008. ``Robotic cell for custom finishing operations.'' \emph{International Journal of Computer Integrated Manufacturing} 21 (1): 33--42.

\bibitem[Nemec and Zlajpah(2008)]{nemec2008shoe}
Nemec, Bojan, and Leon Zlajpah. 2008. ``Shoe Grinding Cell using Virtual Mechanism Approach.'' In \emph{ICINCO-RA (1)}, 159--164.

\bibitem[Oliver et~al.(2021)]{oliver2021towards}
Oliver, Guillermo, Pablo Gil, Jose~F Gomez, and Fernando Torres. 2021. ``Towards footwear manufacturing 4.0: shoe sole robotic grasping in assembling operations.'' \emph{The International Journal of Advanced Manufacturing Technology} 114: 811--827.

\bibitem[Pagano, Russo, and Savino(2020)]{pagano2020vision}
Pagano, Stefano, Riccardo Russo, and Sergio Savino. 2020. ``A vision guided robotic system for flexible gluing process in the footwear industry.'' \emph{Robotics and Computer-Integrated Manufacturing} 65: 101965.

\bibitem[Pedrocchi et~al.(2015)]{pedrocchi2015design}
Pedrocchi, Nicola, Enrico Villagrossi, Claudio Cenati, and Lorenzo Molinari~Tosatti. 2015. ``Design of fuzzy logic controller of industrial robot for roughing the uppers of fashion shoes.'' \emph{The International Journal of Advanced Manufacturing Technology} 77: 939--953.

\bibitem[Rom{\'a}n-Ib{\'a}{\~n}ez, Jimeno-Morenilla, and Pujol-L{\'o}pez(2018)]{roman2018distributed}
Rom{\'a}n-Ib{\'a}{\~n}ez, Vicente, Antonio Jimeno-Morenilla, and Francisco~A Pujol-L{\'o}pez. 2018. ``Distributed monitoring of heterogeneous robotic cells. A proposal for the footwear industry 4.0.'' \emph{International Journal of Computer Integrated Manufacturing} 31 (12): 1205--1219.

\bibitem[Tian et~al.(2016)]{tian2016modeling}
Tian, Fengjie, Chong Lv, Zhenguo Li, and Guangbao Liu. 2016. ``Modeling and control of robotic automatic polishing for curved surfaces.'' \emph{CIRP Journal of Manufacturing Science and Technology} 14: 55--64.

\bibitem[Wang, Dailami, and Matthews(2019)]{wang2019towards}
Wang, Ke~Brian, Farid Dailami, and Jason Matthews. 2019. ``Towards collaborative robotic polishing of mould and die sets.'' \emph{Procedia Manufacturing} 38: 1499--1507.

\end{thebibliography}

\end{document}